\documentclass[11pt]{article}
\usepackage[a4paper,margin=0.95in]{geometry}
\usepackage{amsmath,amssymb,booktabs,array,graphicx}
\usepackage[hidelinks]{hyperref}
\title{\textbf{Remembering Distinct Items, Not Tokens:}\\[3pt]
\textbf{\large A Learnable Dirichlet-Process Cache}\\[1pt]
\textbf{\large Between State-Space Models and Attention}}
\author{Siddharth Pal and Viktoria Rojkova}
\date{Preprint}

\begin{document}
\maketitle

\begin{abstract}
\noindent Fixed-state sequence models such as structured state-space models compress an
unbounded past into a bounded state, which makes them cheap but caps their associative recall
at roughly the state dimension; attention escapes the cap by keeping a key--value entry for
\emph{every} token, at quadratic compute and a cache that grows with the sequence. We study the
middle ground: a sparse cache that allocates a slot only when an input is \emph{novel}, so its
size tracks the number of \emph{distinct} items rather than the number of tokens. The allocation
rule is the DP-means clustering rule, the small-variance limit of a Dirichlet-process mixture,
applied not as latent-variable inference but as the key--value memory
operator for a deep recurrent backbone, a synthesis that, to our knowledge, has not been reported
even though each ingredient (online DP-means clustering, clustered attention, growing caches)
exists in isolation. We develop this cache in two forms --- a static Dirichlet-process cache with a fixed
concentration, and a surprise-adaptive variant whose concentration is adjusted by a running surprise signal that admits capacity during bursts of novel content and reclaims it once the stream
settles. On a controlled associative-recall benchmark with redundancy we show at the mechanism
level that the cache matches full-attention recall while storing only the distinct items
(a fourfold reduction at fourfold redundancy), that it dominates a fixed-budget eviction cache on
the recall-versus-size frontier, and that on a state-space backbone it answers both a recall query
and a long-range aggregate at the lowest memory of any model tested. We further show the
allocation is \emph{learnable end to end}: a two-parameter novelty-threshold gate trained on the
task loss alone recovers the rule exactly, whereas an over-parameterized gate fails, so the
operative ingredient is the inductive bias rather than capacity. We are explicit about scope:
the core evidence is a family of controlled mechanism studies at modest scale --- associative
recall, the recall-versus-size frontier against deployed eviction caches, a learnable-gate
ablation, a state-space hybrid, cost scaling, and the adaptive-concentration test --- run and
cross-checked across CPU and GPU on several machines (Apple M1 and M4, and an NVIDIA T4 on Google
Colab), with the distinct-items property confirmed on four real streams --- recommendation,
systems logs, clinical events, and insurance claims. The cited systems enter as re-implemented
mechanisms rather than full models, and a real-backbone, real-corpus language validation is
pursued in a companion study.
\end{abstract}

\section{The problem, and the gap}

A sequence model summarizes a growing history in a state of fixed size. A structured
state-space model (S4 and its diagonal successors, and the selective recurrence of Mamba) does
this with a bounded state and therefore runs in linear time with constant-size inference; the
price, made precise by a recent line of work, is that such a bounded state cannot perform
content-addressed recall beyond about its own dimension --- the formal copying separation of
Jelassi et al.\ and the empirical recall--throughput trade-off measured by the Zoology
benchmark. Attention removes the cap by retaining a key--value entry for every token and reading
by content, which is why it does recall well, but its cache grows linearly with the sequence and
its compute is quadratic. The two are the endpoints of a single axis: how much state to keep.

Between the endpoints sits a natural idea. If a long stream contains many repeated or redundant
tokens --- as real sequences do --- then the information needed for recall is carried by the
\emph{distinct} items, not by every token, so a memory that opens a new slot only when an input is
genuinely novel, and otherwise merges it into an existing slot, would store on the order of the
number of distinct items and no more. We place this rule in its history rather than claim it whole.
As a clustering primitive it is old: the deterministic ``open a new cluster when the nearest is
farther than a threshold, otherwise join it'' rule is exactly \emph{DP-means} (Kulis and Jordan),
the small-variance, maximum-a-posteriori limit of a Dirichlet-process mixture, and before it the
classical leader or online-clustering rule. What we take from this line is the allocation rule;
what we contribute is its use as the \emph{memory operator itself} for a deep sequence model, where
the cluster set is the key--value cache and its cardinality is the recall cost.

The Dirichlet process has of course already given \emph{classical} sequence models an unbounded
latent cardinality --- the infinite hidden Markov model, the hierarchical Dirichlet process and its
switching linear-dynamical-system variants, and the infinite recurrent switching linear dynamical
system, which grows its discrete mode count by a distance-dependent restaurant process --- but on
shallow, classical backbones and as latent-state inference, not as a key--value memory for a deep
recurrent model.

The nearest neighbours in deep learning each stop one step short of our approach, and
it is worth saying exactly where. \emph{Clustered and routing attention} (Vyas et al.; Roy et al.;
and the locality-sensitive-hashing buckets of Reformer, Kitaev et al.) already merge similar keys to
cheapen attention --- but they cluster \emph{within} a full-attention layer over a bounded context,
as a per-batch grouping whose purpose is to approximate the full softmax, and their cluster count is
a fixed hyperparameter rather than a quantity that grows to the number of distinct items across an
unbounded stream; they also do not position the mechanism against the fixed-state floor or ask
whether the grouping can be learned as an allocation policy. The \emph{growing-memory transformers}
(Memorizing Transformers, the infinite-former, $k$NN language models, neural episodic control, the
Kanerva machine, product-key memories, modern Hopfield networks) grow or address an external memory
by nearest-neighbour retrieval, continuous attention, or a conjugate update, but they keep per-token
entries and do not allocate on novelty. The \emph{fixed-budget sparse caches} (StreamingLLM, H2O,
landmark attention) hold a \emph{constant} budget and \emph{evict} --- by recency or by accumulated
attention mass --- which is the opposite control axis from growing a cache to the number of distinct
items. The \emph{surprise-gated memories} (Titans, EM-LLM) grow on novelty, but through learned
write gates rather than a DP-means allocation. Our contribution is therefore a synthesis rather than
a new primitive: DP-means online allocation used as the cache itself, sized to the distinct items of
an unbounded stream, placed on the state-space-to-attention axis, and shown recoverable end to end by
a minimal learned gate. We have not found this synthesis reported, and we make that novelty claim in
its narrow form deliberately, because each ingredient exists in isolation.

\paragraph{Two approaches.} We develop the middle-ground memory in two forms, which together are
the subject of the paper. \emph{Approach~I, the Dirichlet-process cache}, allocates a slot on
novelty with a \emph{fixed} concentration, so the cache size tracks the number of distinct items;
this is the base construction. \emph{Approach~II, the surprise-adaptive Dirichlet-process cache},
makes the concentration \emph{adaptive}: a surprise signal, akin to a temperature, driven by the
recent allocation rate raises the slot budget during bursts of novel content and lowers it,
consolidating slots, once the stream
becomes predictable. The concentration is thus adjusted by the model's own surprise, admitting
capacity when novel content arrives and reclaiming it once the stream settles, so the cache tracks
non-stationary demand while staying bounded. The two share the same allocate-on-novelty operator and read, and differ only in
whether the concentration is a constant or a temperature-driven schedule.

\paragraph{Contributions.} Around these two approaches we make the following claims, each scoped to
the controlled setting of this study. (i) We identify and position the gap: a Dirichlet-process
allocate-on-novelty sparse memory for a deep recurrent backbone, between a fixed bounded state and
a full attention cache. (ii) The Dirichlet-process cache (Approach~I) stores on the order of the
\emph{distinct} items rather than the tokens, matching full-attention recall at a fraction of the
cache and dominating the deployed fixed-budget eviction caches (H2O, StreamingLLM, SnapKV, and
recency) on the recall-versus-size frontier, on both the synthetic probe and a real heavy-tailed
task. (iii) Composed with a state-space backbone it performs recall and long-range integration
together at the lowest memory of any model tested. (iv) The allocation is learnable end to end from
the task loss by a minimal novelty-threshold gate, so the policy need not be hand-specified, and the
operative ingredient is the inductive bias rather than gate capacity. (v) The distinct-items
property holds unchanged on four real streams from unrelated domains --- recommendation, systems
logs, clinical events, and insurance claims --- with no per-domain tuning of the threshold, and on
system logs the same rule is itself an online log parser. (vi) The surprise-adaptive cache
(Approach~II) beats a fixed budget at equal average cost when demand is non-stationary, and offers
no advantage over a matched fixed budget when demand is stationary, its gain being exactly the
capacity a fixed budget wastes by provisioning for the hardest moment at all times; both sides of
this boundary are confirmed on real streams.

\section{Method}

We present the cache in two forms that share the same allocate-on-novelty operator and read, and
differ only in the concentration: a static \emph{Dirichlet-process cache} (Approach~I), whose
concentration is a fixed threshold, and a \emph{surprise-adaptive} cache (Approach~II), whose
concentration is a temperature-driven schedule that adapts to demand.

\paragraph{Setup, with a worked example.} The probe is associative recall, a controlled synthetic
task chosen to isolate one capability rather than to benchmark a system. A stream presents $K$
distinct \mbox{(key, value)} items, each repeated $r$ times in random order, so the stream has
length $L=rK$ while only $K$ of its tokens are distinct; the model reads the whole stream, is then
given a query key, and must return the value that key was bound to. Keys are vectors and values are
class labels, and the redundancy $r$ is the deliberate lever, because it pulls the number of tokens
$L$ apart from the number of distinct items $K$, which is exactly the quantity a novelty-allocating
cache should track.

Concretely, suppose $K=3$ distinct facts, say (apple, red), (sky, blue), and (grass, green), each
repeated $r=2$ times and shuffled into a stream of $L=6$ tokens such as (sky, blue), (apple, red),
(grass, green), (apple, red), (sky, blue), (grass, green), followed by the query ``apple?'' whose
answer is ``red''. A \emph{slot} is one entry of a model's memory, holding a single key and its
value. Full attention keeps one slot per token and therefore stores all $L=6$ entries; a
fixed-state model keeps no per-item slots at all but folds the stream into one bounded state, so it
cannot keep the facts apart once they outnumber its state dimension; the Dirichlet-process cache
opens a slot only when a token is novel, so it stores the $K=3$ distinct facts and merges each
repeat into the slot it already spawned. The slot count is thus the memory cost we report, and at
redundancy $r$ the cache carries $r$ times fewer entries than attention while answering the same
query. Real streams are of course far longer and far more redundant --- the same word, name, or
identifier recurs many times across a document --- which is the regime this small, transparent probe
is built to expose in isolation.

\paragraph{Approach I: the allocate-on-novelty cache.} The cache is a set of slots, each holding a key and a
value, with a per-slot usage counter. For an incoming pair $(k_t,v_t)$ we measure novelty against
the occupied slots,
\begin{equation}
\mathrm{nov}_t \;=\; 1 - \max_{i \in \text{slots}} \mathrm{sim}(k_t,\,\kappa_i),
\end{equation}
where $\mathrm{sim}$ is cosine similarity and $\kappa_i$ the $i$-th slot key. The Dirichlet-process
rule then decides allocation: if the input is dissimilar to every occupied slot it opens a new
table, otherwise it joins the nearest,
\begin{equation}
\text{if } \mathrm{nov}_t > \tau:\ \text{append } (k_t,v_t)\ \text{(new slot)};\qquad
\text{else}:\ \text{merge into } \arg\max_i \mathrm{sim}(k_t,\kappa_i),
\end{equation}
which is exactly the DP-means rule (the maximum-a-posteriori, small-variance limit of a
Dirichlet-process mixture): a point far from all existing clusters starts a new one with strength
governed by the threshold $\tau$ (playing the role of the concentration), a point close to a cluster
is absorbed. Whether this rule is genuinely a Dirichlet process, and not merely a threshold, is
settled by simulation: run as the underlying stochastic process the slot count grows as
$\alpha\ln N$, with the concentration as the slope (Antoniak, 1974), the Dirichlet-process
signature, whereas on data drawn from a finite set of item types --- the case throughout this
paper --- the same rule instead \emph{saturates} at the true number of distinct types, allocating
exactly as many slots as the data's complexity demands. Because repeats are similar to the slot
they spawned, they merge rather than allocate, so the cache size converges to the number of
distinct items. The slot key is kept at its first occurrence; updating it to the running centroid of
the merged occurrences (online $k$-means) is equivalent at low noise and marginally more stable near
the inter-item separation, so this choice does not change the result. A query $q$ reads by attention over the (few) slots,
\begin{equation}
\hat v \;=\; \sum_i \mathrm{softmax}_i\!\big(\mathrm{sim}(q,\kappa_i)/\theta\big)\,\nu_i,
\end{equation}
decoded to a value class. Read cost is therefore set by the slot count, near $K$, not by $L$.

\paragraph{A learnable gate.} The threshold rule can be replaced by a differentiable gate so the
allocation is learned from the task. We assign each token a keep-probability
$g_t=\sigma\big(a\,(\mathrm{nov}_t-b)\big)$ with learnable scalars $a,b$, where the novelty
feature is computed causally, $\mathrm{nov}_t = 1-\max_{s<t}\mathrm{sim}(k_t,k_s)$. During
training the query attends with a soft mask, adding $\log g_t$ to the attention logits so that a
slot with $g\!\to\!0$ is suppressed and one with $g\!\to\!1$ is kept, which is fully
differentiable; a budget penalty $(\sum_t g_t - M)^2$ pulls the kept count toward the target $M$.
At inference the top-$M$ tokens by $g$ are kept and read as a hard cache. The gate is thus a
learned, amortized version of the allocation rule. The two-parameter gate is one point in a larger
design space, and other parametrizations are possible: a learned similarity metric in place of
cosine, a per-layer or context-dependent threshold produced by a small hypernetwork, a
straight-through or Gumbel-softmax gate that makes a hard keep decision at training time, or a
learned write gate of the kind used by the surprise-gated memories. We test the minimal version
deliberately, to show how little is needed to recover the rule; a fuller comparison of gate
parametrizations is left to future work.

\paragraph{Approach II: adaptive adjustment of the concentration by surprise.} The concentration ($\tau$, or equivalently a slot budget
$M$) is a fixed hyperparameter, but the number of distinct items in play can vary over the stream.
We therefore make it adaptive by driving it with the recent allocation rate. Let
$o_t\in\{0,1\}$ record whether token $t$ opened a new slot ($o_t=1$ when $\mathrm{nov}_t>\tau$, and
$0$ otherwise), and maintain a temperature as its exponential moving average,
\begin{equation}
T_t \;=\; (1-\eta)\,T_{t-1} \;+\; \eta\,o_t ,
\end{equation}
so $T_t$ is high during bursts of novel content and low once the stream is predictable and repeats
dominate. The effective budget then rises with the temperature,
\begin{equation}
M_t \;=\; M_0 + \beta\,T_t ,
\end{equation}
and whenever the slot count exceeds $M_t$ the least-used slot (smallest usage counter, decayed each
step) is evicted:
\begin{equation}
\text{while } |\text{slots}| > M_t:\quad \text{evict } \arg\min_i u_i .
\end{equation}
The cache thus heats during bursts of novel content, admitting capacity, and cools once the stream
settles, reclaiming slots by consolidation, so it keeps a bounded working set that tracks the current
demand. This is an adaptive adjustment of the Dirichlet-process concentration, a schedule that rises
and falls with the data's surprise; the allocation and read rules are otherwise unchanged. We drive the temperature from the allocation rate rather than the raw
per-token novelty because, on a redundant stream, repeats dominate and wash out the brief novelty
spikes, whereas the allocation rate integrates sustained novelty into a clean phase-level signal.

\section{Experiments}

The studies fall in two parts. The first, and larger, part examines \emph{Approach~I}, the static
Dirichlet-process cache with a fixed concentration; the final part examines \emph{Approach~II}, the
surprise-adaptive cache. Each study is presented in the same order --- first the experimental setup,
then the measured result, then the reading it supports. The mechanism studies were run on Apple M1
and Apple M4 machines with fixed seeds, the end-to-end five-model comparison was re-run on an NVIDIA
T4 (Google Colab), and the real-data validation uses public datasets. The cited systems are
re-implemented as minimal mechanisms --- a fixed-state diagonal SSM for the bounded endpoint, full
softmax attention for the unbounded endpoint, a nearest-neighbour cache for the retrieval family, and
recency, heavy-hitter (H2O-style), and sink-plus-window (StreamingLLM-style) caches for the
fixed-budget family --- rather than their full published models, so the comparison is of mechanisms,
not of systems.

\subsection{Approach~I: the static Dirichlet-process cache}

\subsubsection{Recall at the cost of distinct items}
With $K=64$ distinct items at redundancy $r=4$ (so $L=256$ tokens) and sixteen value classes
(chance $0.062$), we compare recall and stored-slot count.
\begin{center}\small
\begin{tabular}{lcc}
\toprule
mechanism (cited family) & recall & slots (cost) \\
\midrule
diagonal SSM, S4D-style & 0.83 & 32 (state dim) \\
full attention & 1.00 & 256 \\
$k$NN cache (store every token) & 1.00 & 256 \\
recency eviction, $B{=}K$ & 0.70 & 64 \\
heavy-hitter eviction, H2O-style, $B{=}K$ & 0.92 & 64 \\
\textbf{DP allocate-on-novelty (ours)} & \textbf{1.00} & \textbf{64} \\
\bottomrule
\end{tabular}
\end{center}
The cache matches full-attention recall while storing one slot per distinct item --- a fourfold
saving at fourfold redundancy --- and beats both a fixed-state SSM, which superposes the stream into
its dimension and degrades (the fixed-state floor's recall depends on task difficulty and state size,
so it varies across the studies below), and every fixed-budget cache at an equal budget, from pure
recency to the heavy-hitter (H2O-style) in the table; we take up those policies and the full
recall-versus-size frontier next. These entries are stable across seeds: rerun over ten seeds of
three hundred episodes each, the cache and full attention hold at $1.000 \pm 0.000$, the heavy-hitter
at $0.92 \pm 0.02$, recency at $0.71 \pm 0.03$, and the fixed-state floor at $0.82 \pm 0.02$.

\begin{figure}[h]\centering
\includegraphics[width=0.92\linewidth]{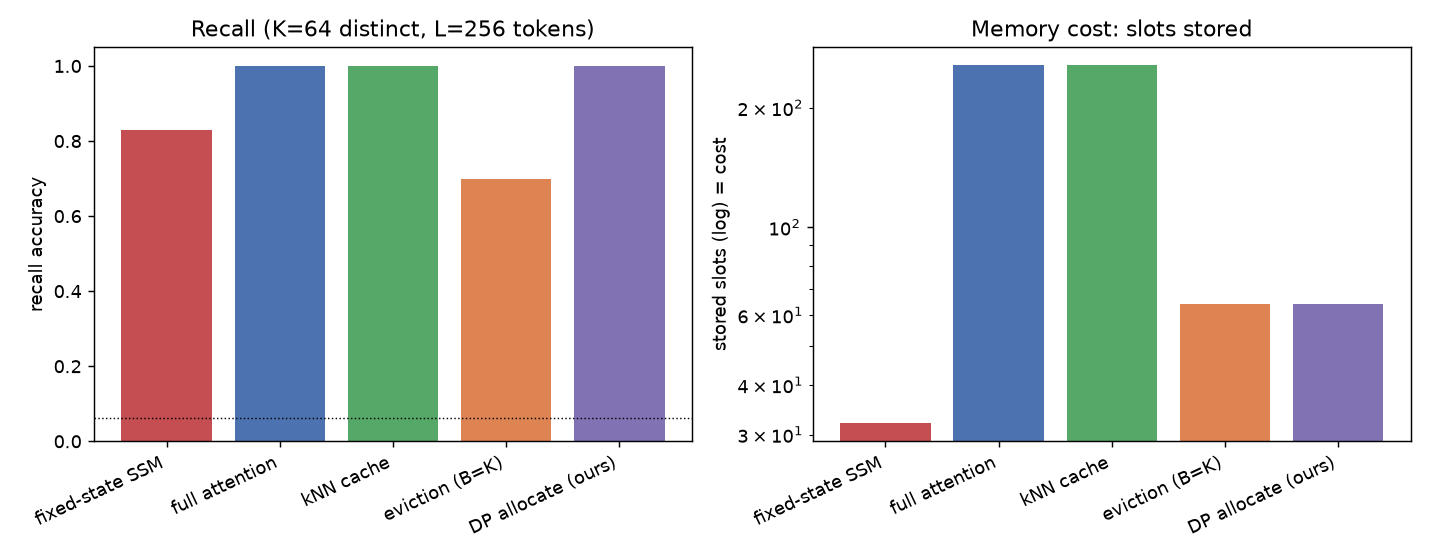}
\caption{Mechanism comparison ($K{=}64$ distinct in $L{=}256$ tokens). Left: recall --- the
DP cache and attention solve the task, the fixed-state SSM and recency eviction fall short. Right:
stored slots (log scale, the cost) --- the DP cache uses one slot per distinct item where
attention and the $k$NN cache store every token.}
\end{figure}

\subsubsection{The concentration knob dominates eviction}
\begin{figure}[h]\centering
\includegraphics[width=0.66\linewidth]{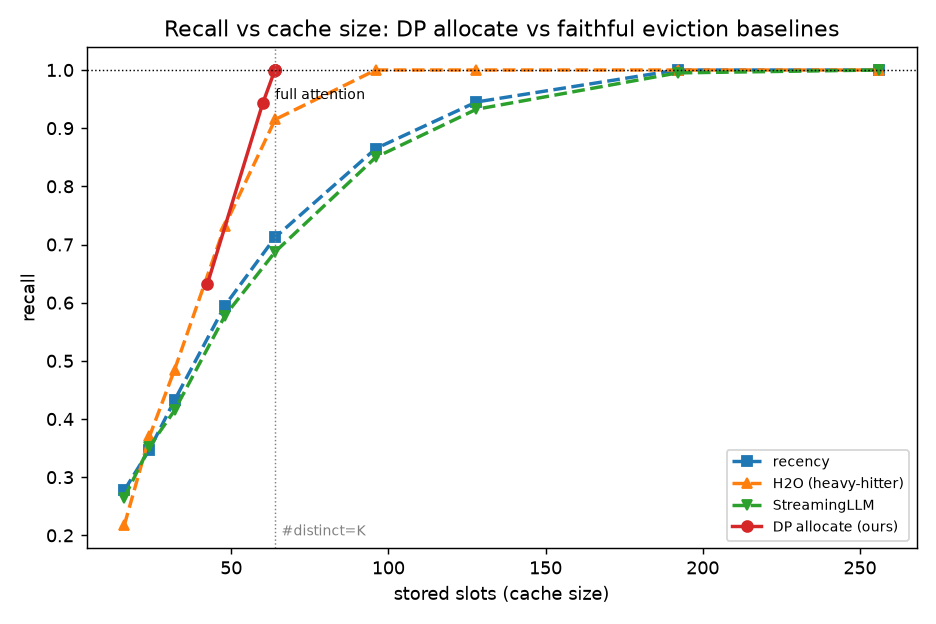}
\caption{\label{fig:faithful}Recall versus cache size against the deployed fixed-budget baselines.
The heavy-hitter (H2O-style) cache, which evicts by accumulated attention mass, is far stronger than
recency or the sink-plus-window (StreamingLLM-style) cache on this redundant stream --- it reaches
$0.92$ at the distinct-items budget --- but the Dirichlet-process allocate-on-novelty frontier
dominates all three, reaching perfect recall at that budget.}
\end{figure}

Sweeping the novelty threshold traces a recall-versus-size frontier for the cache, and sweeping the
budget traces one for each eviction policy. We compare against the deployed policies themselves --- a
heavy-hitter cache (H2O-style), which evicts by accumulated attention mass, and a sink-plus-window
cache (StreamingLLM-style) --- alongside pure recency. As one would expect on a
redundant stream, the heavy-hitter policy is far stronger than recency, reaching $0.92$ at the
distinct-items budget where recency and the sink-plus-window policy reach only about $0.70$; yet the
Dirichlet-process frontier dominates all of them, reaching perfect recall at the distinct-items budget
where every fixed-budget policy still trails, because it keeps exactly the distinct items rather than
approximating them through attention statistics (Figure~\ref{fig:faithful}).

\subsubsection{The frontier holds against the deployed caches on a real task}
\begin{figure}[h]\centering
\includegraphics[width=0.62\linewidth]{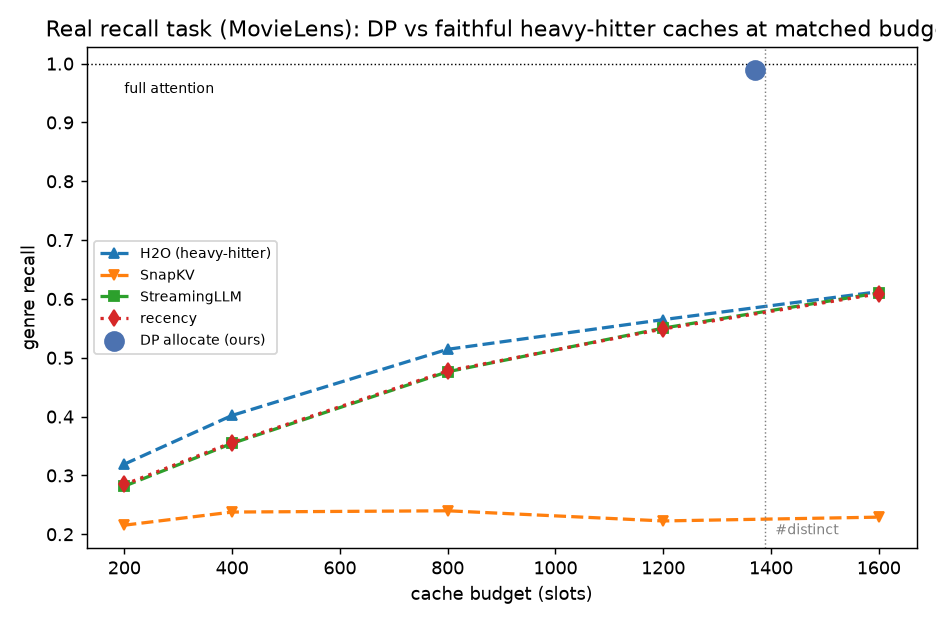}
\caption{\label{fig:realfrontier}Real recall task (MovieLens genre recall), matched budget. The
heavy-hitter (H2O), sink-plus-window (StreamingLLM), recency, and SnapKV caches trail well
below the Dirichlet-process cache, which reaches near-perfect recall at the distinct-items budget by
covering the rare tail the attention-based policies discard.}
\end{figure}

The frontier above is on the synthetic probe; we confirm it on a real, heavy-tailed recall task with
the deployed caches themselves. On the MovieLens stream (thirty thousand occurrences of
$1{,}388$ distinct movies, recall each movie's genre) we run, at a matched budget, a heavy-hitter cache
(H2O-style, evicting by accumulated attention mass), a SnapKV-style cache (keeping the tokens most
attended by a recent observation window), a sink-plus-window cache (StreamingLLM-style), and recency,
against the Dirichlet-process cache and full attention.
\begin{center}\small
\begin{tabular}{lcc}
\toprule
cache on the real task & recall & slots \\
\midrule
full attention & 1.00 & 30{,}000 \\
H2O heavy-hitter & 0.61 & 1{,}600 \\
StreamingLLM / recency & 0.61 & 1{,}600 \\
SnapKV & 0.23 & 1{,}600 \\
\textbf{Dirichlet-process cache} & \textbf{0.99} & \textbf{1{,}370} \\
\bottomrule
\end{tabular}
\end{center}
Even at a budget larger than the number of distinct movies, the strongest fixed-budget policy --- the
heavy hitter --- recalls only $0.61$, because on a heavy-tailed stream it spends its budget on repeated
occurrences of the popular movies and drops the rare tail, which is exactly what the query asks for; the
SnapKV policy, whose window-relevance criterion is tuned for prompt compression rather than covering
distinct entities, does worse still. The Dirichlet-process cache reaches $0.99$ at fewer slots because
it keeps one slot per distinct movie regardless of popularity (Figure~\ref{fig:realfrontier}). The
advantage is precisely rare-tail coverage: a heavy-hitter cache is adequate when only the popular items
are ever queried.

\subsubsection{The allocation is learnable end to end}
We next ask whether the allocation must be hand-specified or can be learned from the task. On
the $K=64$ setting we train gates on the recall loss alone (no rule supervision).
\begin{center}\small
\begin{tabular}{lccc}
\toprule
gate & recall & slots & kept ($g{>}0.5$) \\
\midrule
rule (Dirichlet-process threshold) & 0.999 & 64 & 64 \\
over-parameterized MLP gate & 0.372 & 64 & 0 \\
\textbf{two-parameter novelty gate (ours)} & \textbf{1.000} & 64 & 64 \\
distillation to the rule (upper bound) & 1.000 & 64 & 64 \\
\bottomrule
\end{tabular}
\end{center}
The minimal two-parameter novelty-threshold gate recovers the rule exactly from the task loss,
keeping precisely the distinct items, whereas an over-parameterized gate with the same novelty
feature fails --- it spreads its mass diffusely to satisfy the budget rather than concentrating
on novel tokens. The operative ingredient is therefore the inductive bias (a novelty feature and
just enough parameters to threshold it), not capacity; with it, the Dirichlet-process allocation
is discovered end to end, and distillation, while it also works, is unnecessary. We state this claim
with its boundary: in this probe the repeats of an item share an \emph{identical} embedding, so a
purely per-token saliency score cannot by construction separate one occurrence of an item from
another, and the evidence therefore shows that a novelty feature is \emph{necessary} for a budgeted
selector on this task, not that added capacity is useless in general. A controlled test confirms the failure is not merely
that degeneracy, however: adding per-occurrence noise to break the ties still leaves the
over-parameterized saliency gate near chance, so a novelty feature --- not more capacity --- is what a
budgeted selector needs on this task. Nor is the comparison an artifact of a single training run:
retrained from scratch over five seeds, the two-parameter gate reaches $1.000 \pm 0.000$ while the
over-parameterized gate remains at $0.30 \pm 0.02$.

\subsubsection{On a state-space backbone: recall and integration together}
Placed as the recall module beside a state-space backbone, the cache is tested on a stream that
demands both a recalled value and a long-range aggregate (a running count). The state-space
backbone supplies the aggregate in linear time; the cache supplies the recall at distinct-items
cost. With $K=24$ distinct items in $L=72$ tokens:
\begin{center}\small
\begin{tabular}{lccc}
\toprule
model & recall & integration ($R^2$) & slots \\
\midrule
SSM only & 0.14 & 0.98 & 48 \\
attention only & 1.00 & 0.98 & 72 \\
\textbf{SSM $+$ DP-cache (ours)} & \textbf{1.00} & \textbf{0.99} & \textbf{24} \\
\bottomrule
\end{tabular}
\end{center}
The state-space model alone integrates but cannot recall; attention does both at the token-count
cost; the hybrid does both at the distinct-items cost. This is the middle-ground claim realized
as an architecture rather than only a mechanism.

\begin{figure}[h]\centering
\includegraphics[width=0.92\linewidth]{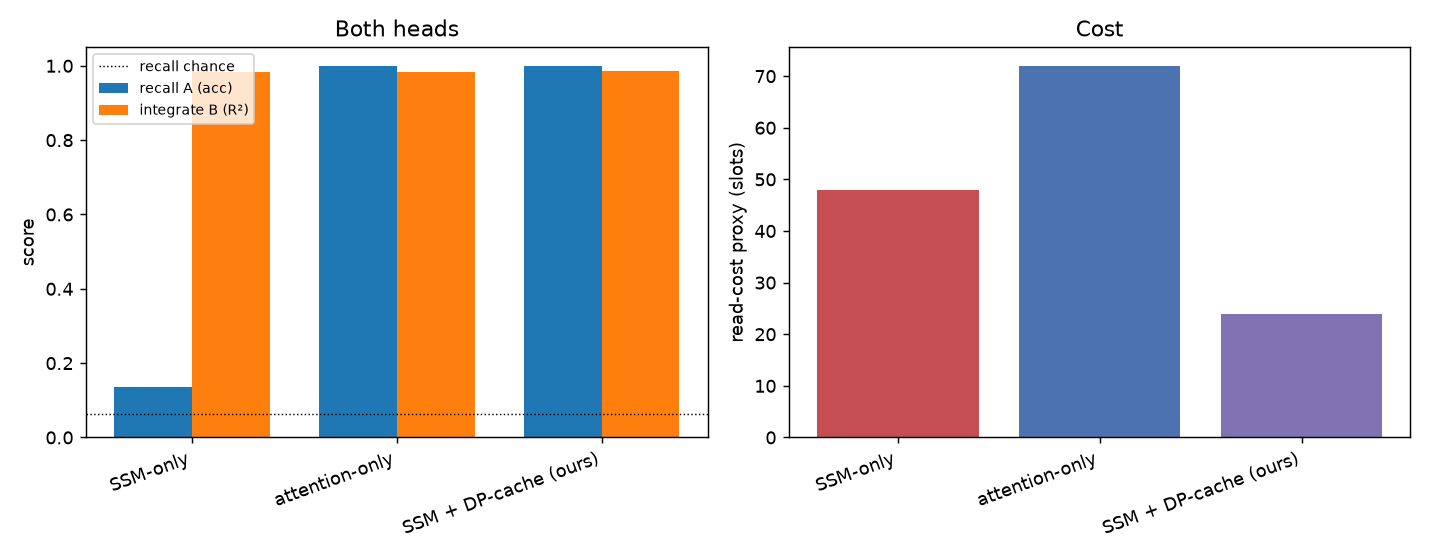}
\caption{State-space backbone with the sparse cache. Left: the hybrid matches attention on both
the recall head and the long-range integration head, where the SSM alone fails recall. Right:
read-cost --- the hybrid uses the fewest slots (the distinct-items count).}
\end{figure}

\subsubsection{End to end on a GPU: the five behaviours in one run}
Scaling to $K=128$ distinct items in $L=512$ tokens with thirty-two value classes (chance $0.031$),
a trained readout, and a real attention baseline, and training every model end to end on a GPU, the
whole story appears in a single run.
\begin{center}\small
\begin{tabular}{lccr}
\toprule
model & recall & slots (cache) & params \\
\midrule
fixed-state SSM (floor) & 0.039 & 128 & 73{,}888 \\
full attention (ceiling) & 1.000 & 512 & 123{,}424 \\
Dirichlet-process cache, rule & 0.999 & 128 & 123{,}424 \\
learned top-$M$ saliency gate & 0.282 & 128 & 140{,}065 \\
\textbf{learned novelty gate (ours)} & \textbf{0.999} & \textbf{128} & 123{,}426 \\
\bottomrule
\end{tabular}
\end{center}
The rule-based cache matches attention's recall ($0.999$ versus $1.000$) at a fourfold smaller
cache, and the fixed-state model ceilings at chance. The learned gate recovers the rule \emph{only} with the
novelty inductive bias: a two-parameter novelty gate reaches attention-level recall at the
distinct-items cache, while an over-parameterized top-$M$ saliency gate with the \emph{same} slot
budget fails, confirming on a GPU what the CPU study showed --- the operative ingredient is the
bias, not the capacity. Note that the cache models carry the \emph{same parameters as attention}
(the novelty selection is parameter-free, or two scalars for the learned gate); the saving is in
the key--value cache, the column of slots, not in the weights.

\subsubsection{Real-data validation across four domains}
To check the central property well beyond the synthetic probe, we run the same allocate-on-novelty
cache, unchanged, on four real redundant streams from unrelated domains --- recommendation, systems
logs, clinical events, and insurance claims --- each a long, heavy-tailed sequence in which the number
of distinct entities is far smaller than the number of events.
\begin{center}\small
\setlength{\tabcolsep}{4pt}
\begin{tabular}{@{}l >{\raggedright\arraybackslash}p{3.3cm} r r c >{\raggedright\arraybackslash}p{3.0cm}@{}}
\toprule
domain & stream & events & \#distinct & redundancy & DP cache (task; slots) \\
\midrule
recommendation & MovieLens-100k, movie occurrences & 100{,}000 & 1{,}682 & $60\times$ & recall $0.99$; $\approx$1{,}670 slots \\
systems logs & Loghub HDFS, log lines & 2{,}000 & 14 & $143\times$ & grouping acc.\ $0.89$; 15 templates \\
clinical & MIMIC-IV demo, prescription events & 18{,}087 & 631 & $29\times$ & recall $0.99$; 628 slots \\
claims & DE-SynPUF, inpatient diagnosis codes & 150{,}000 & 3{,}981 & $38\times$ & recall $0.99$; 3{,}933 slots \\
\bottomrule
\end{tabular}
\end{center}
In each case the cache allocates on the order of the number of distinct entities --- movies, log
templates, drugs --- rather than the number of events, a sixty- to one-hundred-forty-fold reduction,
and solves the domain task at the level of full attention (which stores every event) while a
fixed-budget eviction cache at a quarter of the distinct count falls toward chance. On the
MovieLens ratings, taken chronologically with popularity ranging from $583$ occurrences down to a
handful, the cache recovers each movie's genre at $0.99$ using about $1{,}670$ of $1{,}682$ possible
slots. On the Loghub system-log benchmark the allocate-on-novelty rule \emph{is} an online log parser
--- the mechanism of production parsers such as Drain and Spell --- recovering $15$ templates against
a ground truth of $14$ at a grouping accuracy of $0.89$ and a hundred-forty-fold compression. On the
open MIMIC-IV clinical demo the cache stores $628$ distinct drugs for $18{,}087$ administration events
and recalls each drug's administration route at $0.99$, matching attention's $1.00$ at twenty-nine-fold
fewer slots. And on the public DE-SynPUF synthetic Medicare claims the cache stores $3{,}933$ distinct
diagnosis codes for $150{,}000$ coded events and recalls each code's ICD-9 chapter at $0.99$, a
thirty-eightfold reduction, where the quarter-budget eviction cache again falls sharply (to $0.30$). The
same behaviour holds on the sparser variants (MovieLens genres, MIMIC diagnosis chapters) and degrades
only where a domain is genuinely harder, as on the more diverse BGL log
benchmark, where a bare token similarity trails a tuned parser. We are explicit about what these
establish: for the recommendation and clinical streams the keys are synthetic and the read is
train-free, so they validate the cache \emph{statistics} --- that the allocated set tracks the distinct
entities and supports attention-level recall --- on real, heavy-tailed data rather than an end-to-end
task, whereas the log-parsing result is an end task scored against ground truth. A trained,
real-corpus validation on a real backbone remains the companion study.

\begin{figure}[h]\centering
\includegraphics[width=0.92\linewidth]{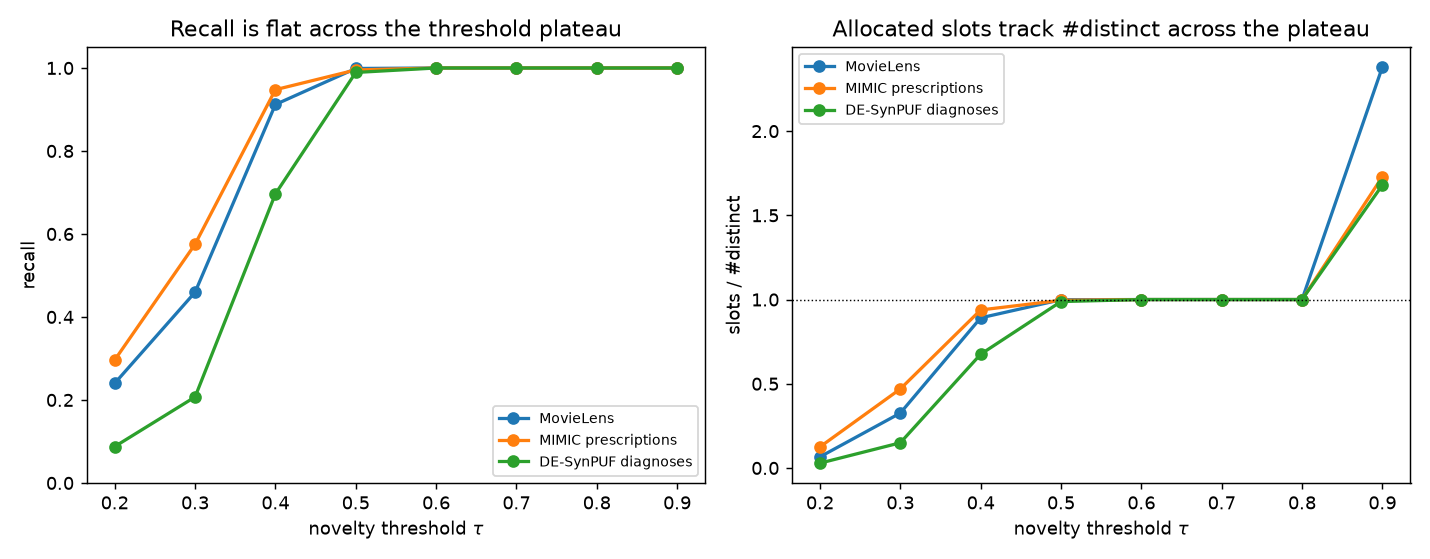}
\caption{\label{fig:tausens}Threshold sensitivity on the real streams. Left: recall is flat at the
ceiling across $\tau \in [0.5, 0.8]$ on all three domains. Right: allocated slots stay pinned at the
distinct count over the same range; below the plateau the cache merges distinct entities, above it
the cache stores benign duplicates.}
\end{figure}

The threshold requires no per-domain tuning. Sweeping $\tau$ from $0.2$ to $0.9$ on the
recommendation, clinical, and claims streams, every value in $[0.5, 0.8]$ yields the same cache on
all three --- the allocated slots equal the distinct count and recall is at least $0.99$ --- because
the threshold only has to fall between two well-separated similarity populations, the near-unit
similarity of a repeat to its own slot and the near-zero similarity between distinct entities. Below
the plateau the cache under-allocates, merging distinct entities and losing recall in proportion; at
$\tau=0.9$, above the repeat similarity at this noise level, it over-allocates benignly, storing
duplicate slots (roughly twice the distinct count) while recall stays perfect
(Figure~\ref{fig:tausens}).

\subsubsection{Online allocation versus offline clustering}
\begin{figure}[h]\centering
\includegraphics[width=0.62\linewidth]{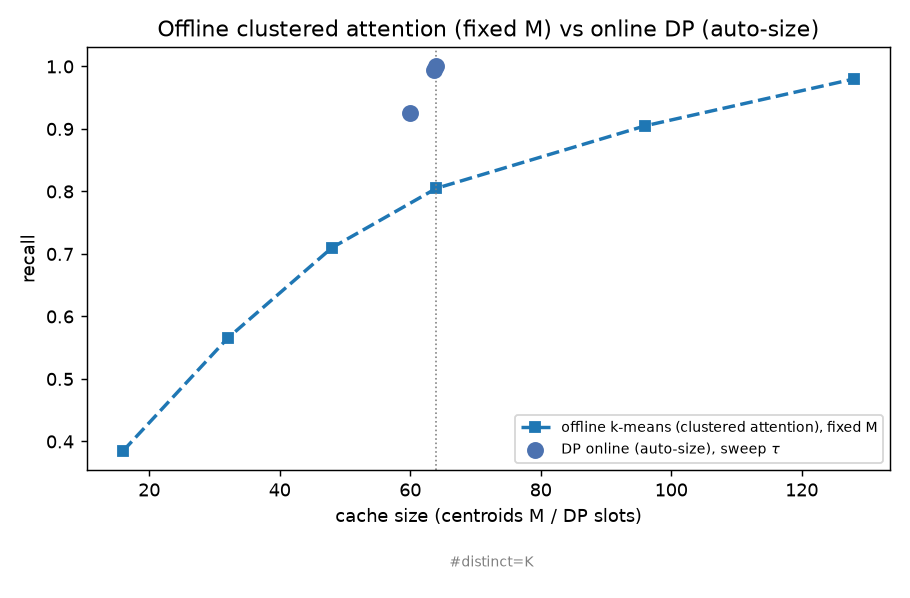}
\caption{\label{fig:cluster}Online allocate-on-novelty versus offline $k$-means (clustered attention).
The offline method must fix the centroid count $M$ and, even at $M=K$, recovers the clusters only
imperfectly; the online DP cache reaches perfect recall at the distinct-items budget by sweeping its
threshold alone, in a single pass and without knowing $K$.}
\end{figure}

Because the allocate-on-novelty rule is DP-means, we contrast it directly with the offline
clustering that clustered attention performs. Running $k$-means over all $L$ keys into a fixed number
of centroids $M$ and reading the query over the centroids traces a recall-versus-$M$ curve that
approaches perfect recall only when $M$ is at least the number of distinct items, and even at $M=K$
Lloyd's algorithm recovers the clusters imperfectly ($0.81$ recall) because it can settle into a local
optimum that merges some distinct items; the online Dirichlet-process cache, sweeping only its
threshold, reaches $1.00$ at the same $64$ slots without being told $K$ and in a single pass
(Figure~\ref{fig:cluster}). The difference is the one the positioning names: offline clustering fixes
the cardinality in advance and needs the whole batch, whereas novelty allocation grows to the distinct
items online.

\subsubsection{Cost}
\begin{figure}[h]\centering
\includegraphics[width=0.92\linewidth]{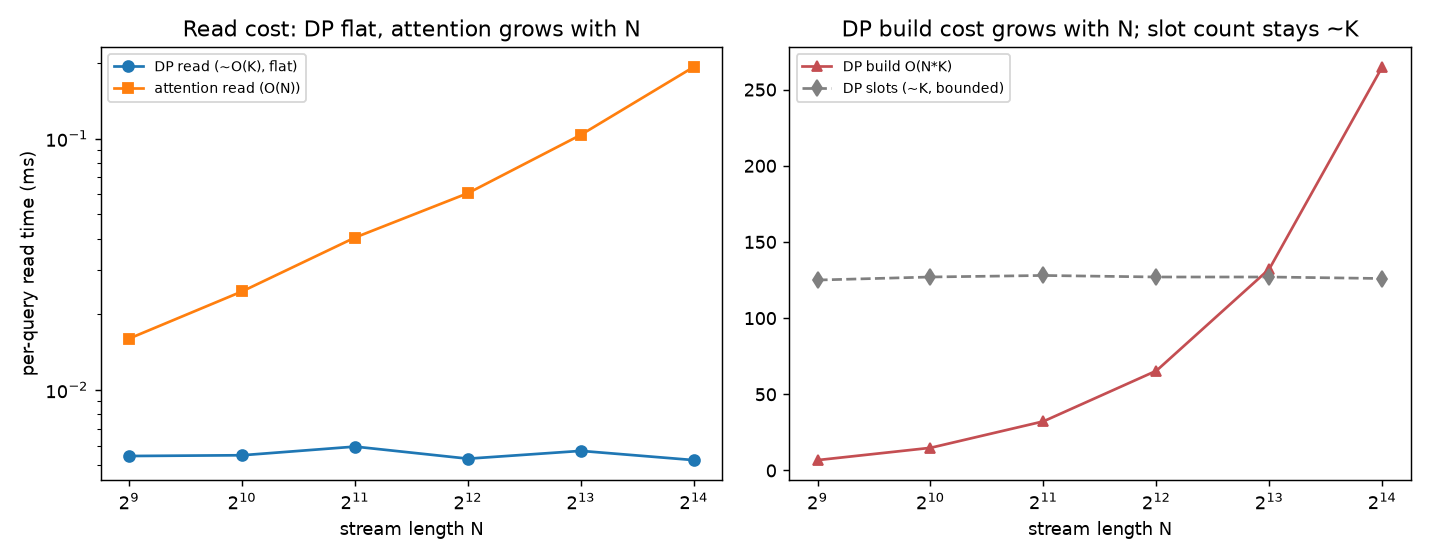}
\caption{\label{fig:cost}Cost versus stream length $N$ at a fixed number of distinct items. Left: the
DP read stays flat ($O(K)$) while attention's read grows ($O(N)$). Right: the slot count stays bounded
near $K$ while the DP build cost grows linearly ($O(N\,|\text{slots}|)$) --- the trade is a heavier
one-time build for cheap reads and bounded memory.}
\end{figure}

The saving is in read and memory, and the allocation is not free. Allocation performs an
$O(|\text{slots}|)$ nearest-slot search per token, so building the cache is $O(N\,|\text{slots}|)$,
heavier than attention's $O(N)$ append; but the cache then holds $\approx K$ slots and a query reads in
$O(K)$ rather than attention's $O(N)$. Measured on the mechanism as the stream length grows at a fixed
number of distinct items, the slot count stays flat near $K$ while attention's per-query read cost
grows with $N$ --- a read speedup from about threefold at redundancy four to thirty-sevenfold at
redundancy one hundred and twenty-eight --- and the build cost grows linearly (Figure~\ref{fig:cost}).
The construction therefore trades a heavier one-time build for cheap repeated reads and bounded memory,
and it is the right trade exactly when reads are many or memory is the binding constraint --- the
regime of long-context inference --- not for a single-pass, read-once workload.

\subsection{Approach~II: the surprise-adaptive cache}

\subsubsection{The surprise-adaptive concentration under non-stationary demand}
Turning to Approach~II, we test the surprise-adaptive concentration on a deliberately non-stationary stream
whose working set alternates between six and thirty items per phase, so the capacity a cache should
hold changes over time. A fixed budget must be set in advance --- either provisioned for the worst
case, paying a permanently larger cache, or matched to the average, in which case it cannot expand
for the hard stretches. The adaptive cache instead lets its budget follow the demand. We bracket the
allocators by the same two endpoints as the static study --- a fixed-state SSM (a bounded
superposition memory) and full attention (one slot per token) --- so that both approaches are read
against the same floor and ceiling.
\begin{center}\small
\begin{tabular}{lcc}
\toprule
allocator & recall & average slots \\
\midrule
fixed-state SSM (floor) & 0.77 & 32 (state dim) \\
full attention (ceiling) & 1.00 & $\sim$450 (grows to 864) \\
fixed-$\tau$ Dirichlet process (unbounded) & 0.99 & 74.4 \\
eviction, budget $=30$ (worst case) & 0.99 & 27.0 \\
eviction, budget $=18$ (matched average) & 0.81 & 16.5 \\
\textbf{adaptive concentration (ours)} & \textbf{0.89} & \textbf{17.5} \\
\bottomrule
\end{tabular}
\end{center}
The endpoints behave as expected under non-stationarity: the fixed-state SSM is stuck at recall
$0.77$ with a bounded state that cannot expand for the hard phases, while full attention reaches
$1.00$ but its cache grows without limit with the tokens (to over eight hundred slots). Between them,
at an equal average cost of roughly seventeen slots the adaptive cache reaches recall $0.89$ where
the matched fixed budget reaches only $0.81$, because its budget swings between about thirteen and
twenty-four in step with the phases --- expanding for the hard stretches and consolidating for the
easy ones --- while staying bounded, unlike the fixed-$\tau$ cache that grows without limit. The boundary is that this advantage is exactly the capacity a fixed budget wastes when it must
provision for the hardest moment at all times: on a \emph{stationary} stream, where demand does not
vary, a fixed budget matched to the working set is just as good, and the adaptive schedule earns its
place only under non-stationary demand.

\begin{figure}[h]\centering
\includegraphics[width=0.92\linewidth]{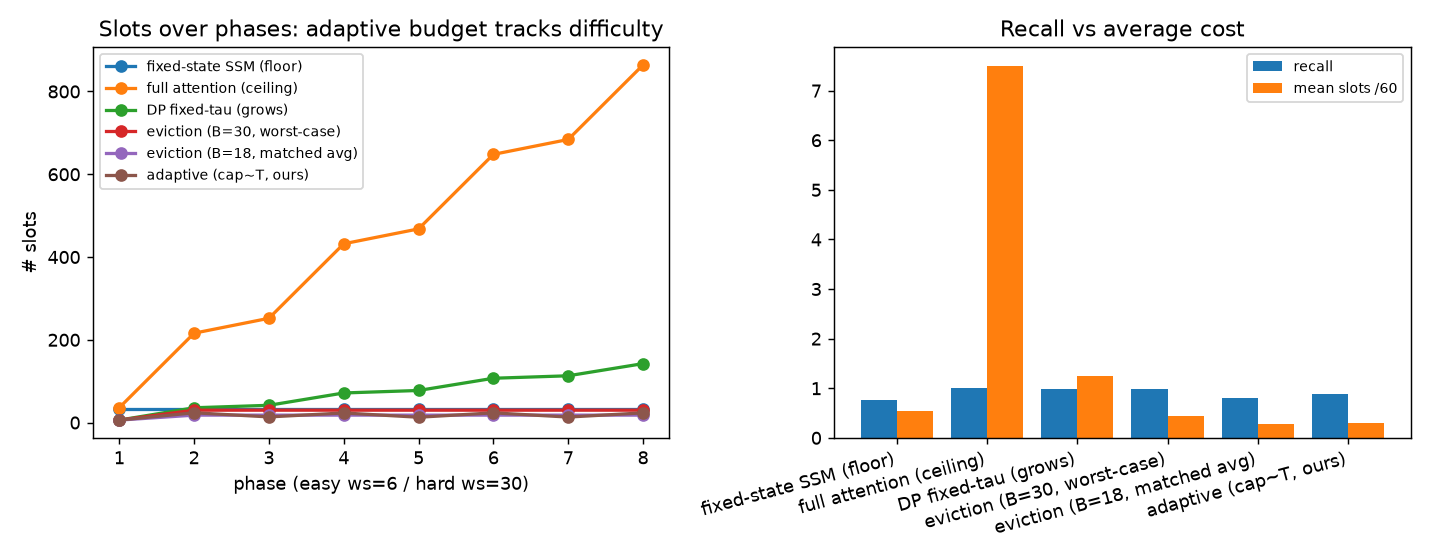}
\caption{Adaptive concentration on non-stationary demand. Left: the budget (slots per phase) swings
with the alternating easy/hard working set, tracking demand while staying bounded, where the
fixed-$\tau$ cache grows without limit and a fixed budget is flat. Right: at equal average cost the
adaptive cache recalls more than a matched fixed budget.}
\end{figure}

\begin{figure}[h]\centering
\includegraphics[width=0.92\linewidth]{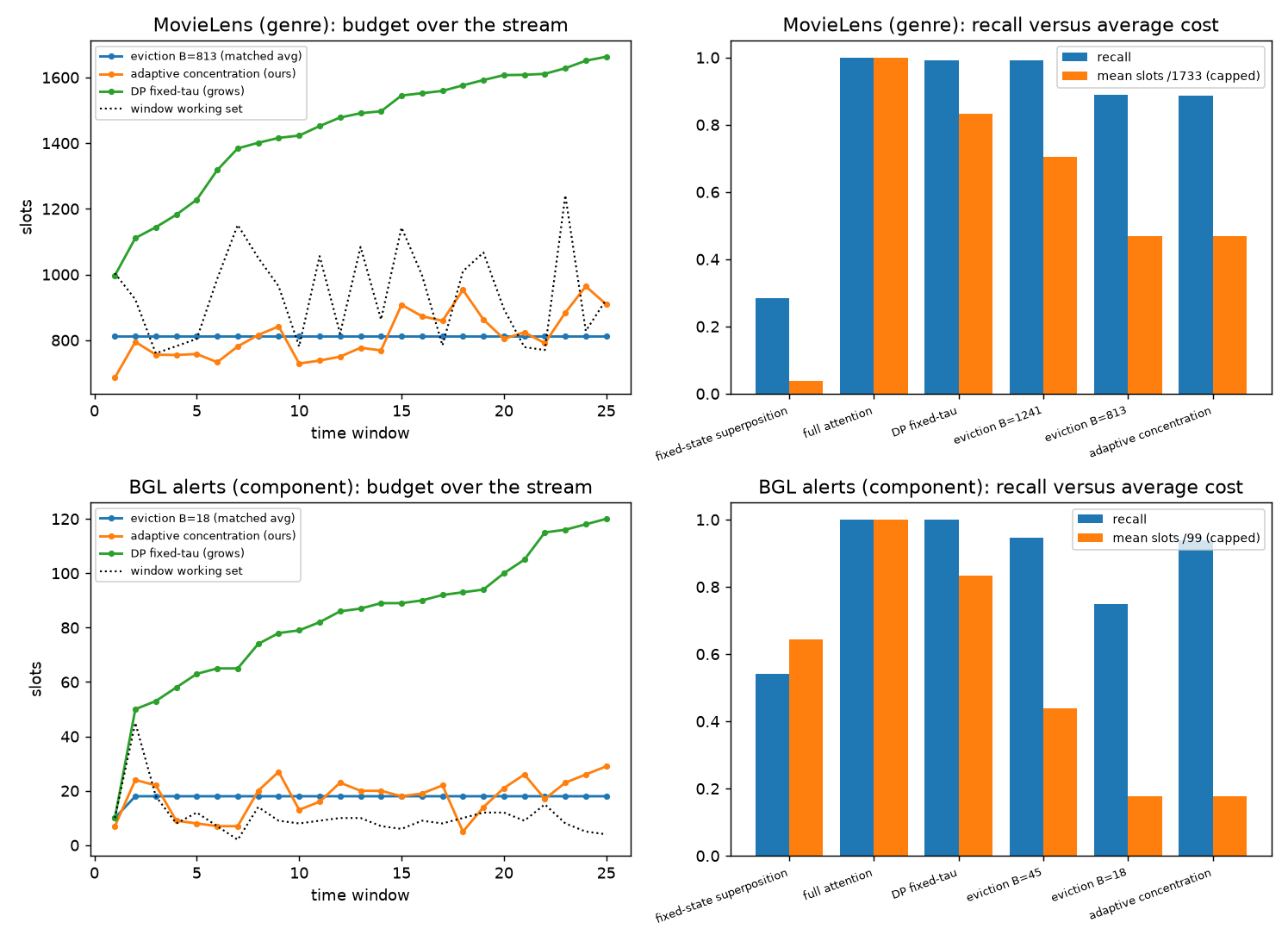}
\caption{\label{fig:adaptreal}The surprise-adaptive concentration on two real streams. Top
(MovieLens): the per-window demand is nearly stationary, and the adaptive budget and a matched
fixed budget tie. Bottom (BGL alerts): the demand is bursty; the adaptive budget tracks the storms
(left) and, at equal average slots, out-recalls the matched fixed budget and approaches the
worst-case-provisioned one (right).}
\end{figure}

Both sides of this boundary hold on real streams. On the MovieLens stream cut into twenty-five
equal calendar-time windows, with each window's distinct movies queried at its close, the
per-window working set turns out nearly flat (it varies by a factor of $1.6$), and the adaptive
budget ties a matched fixed budget --- $0.887$ against $0.890$ at $813$ average slots --- so the
stationary side behaves exactly as stated. On the BGL supercomputer log, whose alert templates
arrive in storms and whose per-window working set swings from two to forty-five distinct templates,
the adaptive cache at the same eighteen average slots recalls $0.94$ where the matched fixed budget
recalls $0.75$, approaching the recall of a worst-case-provisioned budget ($0.95$) at two and a
half times fewer average slots (Figure~\ref{fig:adaptreal}).

\section{Scope, limitations, and future work}

We state the boundaries as part of the contribution rather than as a coda. The evidence is on a
controlled, synthetic associative-recall task at modest scale, verified on a single CPU and re-run
end to end on a GPU; the keys are
well-separated by construction, so the novelty signal is clean, and on real, overlapping data the
threshold and the merge step would need care. The baselines are re-implemented mechanisms, not
the full Mamba, Memorizing-Transformer, or StreamingLLM systems, so the comparison establishes
the behaviour of the mechanisms, not a head-to-head of systems. The contribution we claim is
correspondingly scoped: a Dirichlet-process allocate-on-novelty sparse cache is a coherent,
learnable middle ground between fixed-state recurrence and full attention, it stores on the order
of the distinct items, it dominates fixed-budget eviction on the recall-versus-size frontier, it
composes with a state-space backbone to do recall and integration together, and its allocation is
recoverable from the task by a minimal learned gate. The natural and necessary next step, which
the present CPU study cannot reach, is a real-backbone, real-corpus validation: a Mamba backbone
with the sparse cache against Mamba, full attention, and a fixed-window hybrid, on a long-context
language task where recall matters --- character- or word-level language modelling on a corpus such
as enwik8, WikiText-103, or PG-19, or a needle-in-context retrieval suite such as Long Range Arena
or RULER --- measuring recall, perplexity, cache size, and latency at scale. A first-stage GPU
derisk of the mechanism (the five-model comparison above, trained end to end on a GPU) and
real-stream checks of its statistics across four domains --- recommendation, systems logs, clinical
events, and insurance claims, above --- have now been run and confirm the synthetic findings, as do
the seed-variance re-measurement of the headline tables, the threshold sweep across three of the
real domains, and the two-sided real-stream test of the adaptive concentration. The
full real-backbone, real-corpus validation --- a coupled state-space-plus-sparse-cache language model
against a Mamba backbone, full attention, and a fixed-window hybrid --- is the subject of a companion
study; the present paper establishes the mechanism and that its distinct-items property holds across
synthetic and real streams.

\section*{Acknowledgements}
The mechanism and real-stream studies were designed, run, and verified on an Apple M1 and an
Apple M4, and we are grateful to Google Colab for access to the NVIDIA T4 GPUs, on which the
end-to-end GPU comparison was run. We thank Aarav Pal for his help with the data downloads and
with testing and running the experiments.


\begin{thebibliography}{Ramsauer et~al.(2021)}\small

\bibitem[Antoniak(1974)]{antoniak} C.~E.~Antoniak. Mixtures of Dirichlet Processes with Applications to Bayesian Nonparametric Problems. \emph{The Annals of Statistics}, 2(6):1152--1174, 1974.
\bibitem[Arora et~al.(2023)]{zoology} S.~Arora, S.~Eyuboglu, A.~Timalsina, I.~Johnson, M.~Poli, J.~Zou, A.~Rudra, and C.~R\'e. Zoology: Measuring and Improving Recall in Efficient Language Models. 2023. arXiv:2312.04927.
\bibitem[Beal et~al.(2002)]{ihmm} M.~J.~Beal, Z.~Ghahramani, and C.~E.~Rasmussen. The Infinite Hidden Markov Model. \emph{NeurIPS}, 2002.
\bibitem[Behrouz et~al.(2024)]{titans} A.~Behrouz, P.~Zhong, and V.~Mirrokni. Titans: Learning to Memorize at Test Time. 2024. arXiv:2501.00663.
\bibitem[Du \& Li(2016)]{spell} M.~Du and F.~Li. Spell: Streaming Parsing of System Event Logs. \emph{ICDM}, 2016.
\bibitem[Fountas et~al.(2025)]{emllm} Z.~Fountas et al. Human-inspired Episodic Memory for Infinite Context LLMs (EM-LLM). \emph{ICLR}, 2025. arXiv:2407.09450.
\bibitem[Fox et~al.(2011)]{hdpslds} E.~B.~Fox, E.~B.~Sudderth, M.~I.~Jordan, and A.~S.~Willsky. Bayesian Nonparametric Inference of Switching Linear Dynamical Systems. \emph{IEEE Transactions on Signal Processing}, 59(4):1569--1585, 2011.
\bibitem[Geadah et~al.(2024)]{irslds} V.~Geadah, International Brain Laboratory, and J.~W.~Pillow. Parsing Neural Dynamics with Infinite Recurrent Switching Linear Dynamical Systems. \emph{ICLR}, 2024.
\bibitem[Gu et~al.(2022a)]{s4} A.~Gu, K.~Goel, and C.~R\'e. Efficiently Modeling Long Sequences with Structured State Spaces (S4). \emph{ICLR}, 2022. arXiv:2111.00396.
\bibitem[Gu et~al.(2022b)]{s4d} A.~Gu, A.~Gupta, K.~Goel, and C.~R\'e. On the Parameterization and Initialization of Diagonal State Space Models (S4D). \emph{NeurIPS}, 2022. arXiv:2206.11893.
\bibitem[Gu \& Dao(2023)]{mamba} A.~Gu and T.~Dao. Mamba: Linear-Time Sequence Modeling with Selective State Spaces. 2023. arXiv:2312.00752.
\bibitem[He et~al.(2017)]{drain} P.~He, J.~Zhu, Z.~Zheng, and M.~R.~Lyu. Drain: An Online Log Parsing Approach with Fixed Depth Tree. \emph{ICWS}, 2017.
\bibitem[Jelassi et~al.(2024)]{repeat} S.~Jelassi, D.~Brandfonbrener, S.~M.~Kakade, and E.~Malach. Repeat After Me: Transformers are Better than State Space Models at Copying. \emph{ICML}, 2024. arXiv:2402.01032.
\bibitem[Khandelwal et~al.(2020)]{knnlm} U.~Khandelwal, O.~Levy, D.~Jurafsky, L.~Zettlemoyer, and M.~Lewis. Generalization through Memorization: Nearest Neighbor Language Models. \emph{ICLR}, 2020. arXiv:1911.00172.
\bibitem[Kitaev et~al.(2020)]{reformer} N.~Kitaev, {\L}.~Kaiser, and A.~Levskaya. Reformer: The Efficient Transformer. \emph{ICLR}, 2020. arXiv:2001.04451.
\bibitem[Kulis \& Jordan(2012)]{dpmeans} B.~Kulis and M.~I.~Jordan. Revisiting k-means: New Algorithms via Bayesian Nonparametrics. \emph{ICML}, 2012. arXiv:1111.0352.
\bibitem[Lample et~al.(2019)]{pkm} G.~Lample, A.~Sablayrolles, M.~Ranzato, L.~Denoyer, and H.~J\'egou. Large Memory Layers with Product Keys. \emph{NeurIPS}, 2019. arXiv:1907.05242.
\bibitem[Li et~al.(2024)]{snapkv} Y.~Li, Y.~Huang, B.~Yang, B.~Venkitesh, A.~Locatelli, H.~Ye, T.~Cai, P.~Lewis, and D.~Chen. SnapKV: LLM Knows What You are Looking for Before Generation. \emph{NeurIPS}, 2024. arXiv:2404.14469.
\bibitem[Martins et~al.(2022)]{infty} P.~H.~Martins, Z.~Marinho, and A.~F.~T.~Martins. $\infty$-former: Infinite Memory Transformer. \emph{ACL}, 2022. arXiv:2109.00301.
\bibitem[Mohtashami \& Jaggi(2023)]{landmark} A.~Mohtashami and M.~Jaggi. Landmark Attention: Random-Access Infinite Context Length for Transformers. \emph{NeurIPS}, 2023. arXiv:2305.16300.
\bibitem[Pritzel et~al.(2017)]{nec} A.~Pritzel, B.~Uria, S.~Srinivasan, A.~Puigdom\`enech, O.~Vinyals, D.~Hassabis, D.~Wierstra, and C.~Blundell. Neural Episodic Control. \emph{ICML}, 2017. arXiv:1703.01988.
\bibitem[Ramsauer et~al.(2021)]{hopfield} H.~Ramsauer et al. Hopfield Networks is All You Need. \emph{ICLR}, 2021. arXiv:2008.02217.
\bibitem[Roy et~al.(2021)]{routing} A.~Roy, M.~Saffar, A.~Vaswani, and D.~Grangier. Efficient Content-Based Sparse Attention with Routing Transformers. \emph{TACL}, 2021. arXiv:2003.05997.
\bibitem[Teh et~al.(2006)]{hdp} Y.~W.~Teh, M.~I.~Jordan, M.~J.~Beal, and D.~M.~Blei. Hierarchical Dirichlet Processes. \emph{Journal of the American Statistical Association}, 101(476):1566--1581, 2006.
\bibitem[Vyas et~al.(2020)]{clustered} A.~Vyas, A.~Katharopoulos, and F.~Fleuret. Fast Transformers with Clustered Attention. \emph{NeurIPS}, 2020. arXiv:2007.04825.
\bibitem[Wu et~al.(2018)]{kanerva} Y.~Wu, G.~Wayne, A.~Graves, and T.~Lillicrap. The Kanerva Machine: A Generative Distributed Memory. \emph{ICLR}, 2018. arXiv:1804.01756.
\bibitem[Wu et~al.(2022)]{memtransformer} Y.~Wu, M.~N.~Rabe, D.~Hutchins, and C.~Szegedy. Memorizing Transformers. \emph{ICLR}, 2022. arXiv:2203.08913.
\bibitem[Xiao et~al.(2024)]{streamingllm} G.~Xiao, Y.~Tian, B.~Chen, S.~Han, and M.~Lewis. Efficient Streaming Language Models with Attention Sinks (StreamingLLM). \emph{ICLR}, 2024. arXiv:2309.17453.
\bibitem[Zhang et~al.(2023)]{h2o} Z.~Zhang et al. H2O: Heavy-Hitter Oracle for Efficient Generative Inference of Large Language Models. \emph{NeurIPS}, 2023. arXiv:2306.14048.

\end{thebibliography}
\end{document}